\documentclass[conference,onecolumn]{IEEEtran}

\usepackage{amsmath,graphicx}
\usepackage[ruled]{algorithm2e}
\usepackage{threeparttable}
\usepackage{dcolumn}
\usepackage{booktabs}
\usepackage{amsmath,graphicx}
\usepackage{url}
\usepackage{cite}
\usepackage{psfrag}
\usepackage{amssymb}
\usepackage{amsbsy}
\usepackage{graphicx}
\usepackage{array}
\usepackage{multirow}
\usepackage{bm}
\usepackage{subfigure}
\usepackage{cite}
\usepackage{verbatim}
\usepackage{colortbl}
\usepackage{algorithmic}
\newcommand{\xx}{{\bf x}}

\ifCLASSINFOpdf

\else

\fi

\begin{document}

\title{Image Super-Resolution Based on Sparsity Prior via Smoothed $l_0$ Norm}

\author{\IEEEauthorblockN{Mohammad Rostami}
\IEEEauthorblockA{Department of Electrical and\\Computer Engineering\\
University of Waterloo\\
Waterloo, Canada } \and \IEEEauthorblockN{Zhou Wang}
\IEEEauthorblockA{Department of Electrical and\\Computer Engineering\\
University of Waterloo\\
Waterloo, Canada} }

\maketitle

\begin{abstract}
In this paper we aim to tackle the problem of reconstructing a
high-resolution image from a single low-resolution input image,
known as single image super-resolution. In the literature, sparse
representation has been used to address this problem, where it is
assumed that both low-resolution and high-resolution images share
the same sparse representation over a pair of coupled jointly
trained dictionaries. This assumption enables us to use the
compressed sensing theory to find the jointly sparse
representation via the low-resolution image and then use it to
recover the high-resolution image.  However, sparse representation of a
signal over a known dictionary is an ill-posed, combinatorial
optimization problem. Here we propose an algorithm that adopts the smoothed $l_0$-norm (SL0)
approach to find the jointly sparse representation. Improved
quality of the reconstructed image is obtained for most images in
terms of both peak signal-to-noise-ratio (PSNR) and structural similarity (SSIM) measures.

\end{abstract}


\IEEEpeerreviewmaketitle

\section{INTRODUCTION}
Image super-resolution provides a low cost, software-based technique to improve
the spatial resolution of an image beyond the limitations of the imaging hardware
devices. The areas of application include medical imaging
\cite{101} and satellite imaging \cite{102} and high-definition television (HDTV). In such cases, it is
standard to assume that the observed low-resolution image(s) is a
blurred and downsampled version of the high-resolution image. The
goal is then to recover the high-resolution image using its low-resolution observation(s). Meanwhile, with the growing
capabilities of high-resolution displays, effective image
super-resolution algorithms is essential for us to make the best use of such
devices.

The most typical super-resolution methods require multiple
low-resolution images, with sub-pixel accuracy alignment. In this
approach super-resolution can be considered as an inverse
problem, where it is essential to assume a prior on the solution
to regularize the ill-posed nature of the problem and
avoid infinitely many solutions \cite{103,104}.

In some applications, the number of low-resolution images is limited
and it is desired to reconstruct the high resolution image  from
a single low-resolution image. One approach to overcome this
limitation is to use interpolation methods \cite{106,107,108},
where  unknown pixels of the high-resolution image are estimated
using nearby known pixels from the low-resolution image based on
some assumptions on the relation between pixels. Simple
interpolation methods such as bilinear and bicubic interpolations
result in blued images with ringing artifacts near  edges.
While more advanced methods \cite{107,108} try to avoid this
problem by exploiting natural image priors, they are often
limited in accounting for the complexity of images which have
regions with fine textures or smooth shadings. As a result,
watercolor-like artifacts are often observed in some regions.

A more successful class of methods for single input image
super-resolution are learning based methods. Here a co-occurrence
prior between the high-resolution and low-resolution images is
used to reconstruct the high-resolution image
\cite{109,110,111,30}. The learning procedure has been done via
different schemes such as Markov Random Field (MRF), Primal
Sketch and Locally Linear Embedding (LLE) \cite{109,110,111}.
These algorithms require enormous databases to handle the learning
stage and thus are computationally expensive. One of the recently developed
successful methods \cite{30} uses sparsity
as the co-occurrence prior between the high-resolution and
low-resolution images. This approach reduces the size of the
training database and consequently the computational load. In this
method it is assumed that both images share the same sparse
representation over a pair of jointly trained dictionaries. Using
this assumption we can find the jointly sparse representation via
the low-resolution image and then use it to recover the
high-resolution image. Finding the jointly sparse representation
properly, is important and influences the quality of the
reconstruction result. In the current paper our intention is to
improve the performance of this algorithm  to find the jointly
sparse representation more accurately and thus to improve the
quality of the reconstructed high-resolution image.

Sparse representation of a vector over a known dictionary  is an
ill-posed, combinatorial optimization problem \cite{4}. Several
relaxation approaches have been used to convexify this problem.
The most common approaches are Matching Pursuit (MP)\cite{114},
Basis Pursuit (BP) \cite{45}, and Focal Underdetermined System
Solution (FOCUSS) \cite{115}. In this paper, we adopt a recently proposed Smoothed $l_0$-norm (SL)) algorithm \cite{112}, a faster
solver  for sparse representation, to lessen the
computational load and to improve the quality of the
reconstructed high-resolution image.

The rest of the paper is organized as follows. In section II, we
briefly present the approach that is used to decompose a signal
on an overcomplete dictionary. Section III is devoted to the
super-resolution algorithm and its alternation. Section VI
summarizes experimental results. The paper is finally concluded
in section V.

\section{SPARSE REPRESENTATION VIA SMOOTHED $l_0$ NORM}
Finding the sparsest representation of a source signal over an
overcomplete dictionary is formulated under the topic of
compressed sensing (CS) theory \cite{4,13}. Let $\xx \in
\mathbb{R}^n$ be the source signal, and $D \in \mathbb{R}^{m \times
n}$ the dictionary ($n>m$). Our goal is to find the sparsest
solution of the following linear system:
\begin{equation} \label{1}
\xx = D \alpha,
\end{equation}
where $\alpha \in \mathbb(R)^m$. The recovery of $\alpha$ from
$\xx$ based on \eqref{1} is impossible to implement in a unique
and stable way, unless it is known that $\alpha$ is sparse enough
to have a relatively low value of $\|\alpha\|_0$ \cite{4}.
This assumption is only valid if the dictionary is chosen or
learned properly. Several algorithms have been provided to design
such dictionaries \cite{116,117,118}.

Equivalently \eqref{1} can be formulated as the following
optimization problem:
\begin{equation} \label{2}
(P_0)\hspace{5mm}\min\|\alpha\|_0,\hspace{3mm} \text{s.t.}
\hspace{3mm} \xx=D\alpha
\end{equation}
Due to the highly nonconvex nature of $l_0$-norm, this problem is
ill-posed and intractable. In conventional CS theory \cite{4,13}
it is proven that if the dictionary $D$ satisfies the restricted
isometry property (RIP) \cite{4, 13} with respect to a certain
class of sparse signals to which $\alpha$ is assumed to belong,
then  $\alpha$ can be recovered as a solution to \cite{20, 21}
\begin{equation} \label{3}
(P_1)\hspace{5mm}\min\|\alpha\|_1,\hspace{3mm} \text{s.t.}
\hspace{3mm} \xx=D\alpha
\end{equation}
which is a convex minimization problem. It is straightforward to
reformulate this equivalent problem in terms of linear
programming. This approach results in a tractable problem but is
still time-consuming for large scale systems.  It is also
important to note that the equivalence between $l_0$-norm and
$l_1$-norm is only valid asymptotically and does not always hold
\cite{113}. In a different approach, the $l_0$-norm is approximated directly by a smooth convex
function \cite{112}. This approach has proved to be faster with
possibility of resulting in sparser representation \cite{112}.

Consider the smooth function, $f_{\sigma}(x)=e^{-x^2/\sigma^2}$.
As $\sigma$ approaches zero, we have the following equivalence:
\begin{equation} \label{4}
\|\xx\|_0=n-\sum_{i=0}^n \lim_{\sigma \to
0^+}f_{\sigma}(x_i)=n-\sum_{i=0}^n f_{0^+}(x_i)
\end{equation}
This equivalence does not help us in practice. However, one can
assume that if $\sigma$ is set to be nonzero and sufficiently
small then we can approximate the $l_0$-norm of a vector by
\begin{equation} \label{5}
\|\xx\|_0\approx n-\sum_{i=0}^n f_{\sigma}(x_i)
\end{equation}
This enables us to approximate the $l_0$-norm with a smooth,
differentiable function. This is the key fact that enables us to
replace the $l_0$-norm minimization with a convex problem, so
that we can take advantage of common techniques, such as steepest
descent, to tackle the optimization problem. The value of $\sigma$
controls the trade-off between the closeness to the $l_0$-norm and the
smoothness of the approximation. Now if we define
$F_{\sigma}(\xx)=\sum_{i=0}^n f_{\sigma}(x_i)$, then the minimization
of $\|\xx\|_0$ can be done by maximizing $F_{\sigma}$ by
choosing a proper value for $\sigma$. Due to the nonconvex nature
of the $l_0$-norm, $F_{\sigma}$ will archive a lot of local extreme
points for small values of $\sigma$. Consequently, finding the
global maxima will become difficult. On the other hand, if the
value of $\sigma$ is chosen to be sufficiently large, there will
be no local maxima \cite{112} and asymptotically the solution for
$\sigma=\infty$ is equivalent to the $l_2$-norm solution. Considering
these facts, the authors of \cite{112} provided Algorithm 1 to solve the
optimization problem $P_0$.
\begin{algorithm}
\setlength{\leftmargini}{0pt} \caption{Sparse Representation
using smoothed $l_0$-norm}
\begin{enumerate}
\item {\it Initialization:} \\a. Choose an arbitrary solution from the feasible set, e.g., the minimum $l_2$-norm solution of $D\alpha=\xx$, i.e., $\alpha=D^T (DD^T)^{-1}\xx$.\\b. Choose a decreasing sequence for $\sigma$, $[\sigma_1,...,\sigma_K]$.
\item {\it for $k=1,...,K$:}\\a. Let $\sigma=\sigma_k$.\\b. Maximize the function $F_{\sigma}$ using $L$ iterations
of the steepest ascent algorithm and call it $v_k$.
\item {\it Final answer is $\alpha=v_K$.}
\end{enumerate}
\label{algo1}
\end{algorithm}
Here, the final estimation of each step is used for the
initialization of the next steepest ascent. By a proper selection
of the sequence of $\sigma$, we may avoid being trapped in
the local maxima. Compared to conventional CS solvers, this
algorithms proves to be faster with the possibility of recovering a
sparser solution \cite{112}.
\section{IMAGE SUPER-RESOLUTION BASED ON SPARSE REPRESENTATION}
Sparse representation has been applied to multiple inverse
problems in image processing such as denoising \cite{28},
restoration \cite{29} and super-resolution \cite{30,104}.
Generally sparsity is used as a prior on source signal to avoid
ill-posed nature of inverse problems. In such applications, there
exist a stage, which involves expansion of a source signal over
an overcomplete dictionary, sparsely. The output quality of these
algorithms depends on the accuracy in finding the sparse
representation. In this section, SL0 algorithm is employed in a
super-resolution algorithm  to improve quality of the output high-resolution
image.

Assume that the low-resolution image $\mathbf{Y}$ is produced 
from a high resolution image $\mathbf{X}$ by
\begin{equation}
\mathbf{Y} = SH\mathbf{X} \label{eq:SRprob}
\end{equation}
where \(H\) represents a blurring matrix, and \(S\) is a
downsampling matrix. The recovered high-resolution output image
must be consistent with the low-resolution input image. This
problem is highly ill-posed and infinitely many solutions satisfy
\eqref{eq:SRprob} and are consistent with low-resolution image. To
provide a unique solution, local sparsity model maybe applied as a
prior. We assume that there exist two dictionaries, $D_l$ and
$D_h$, for which each patch of low, $y$, and high, $x$,
resolution images can be represented sparsely simultaneously and
jointly as follows:
\begin{equation}
\begin{split}
x=D_h\alpha\\
y=D_l\alpha
\end{split}
\label{6}
\end{equation}
 These coupled dictionaries are trained simultaneously and jointly over a set of low/high
resolution images such that both low/high resolution images
result in the same sparse representation coefficients. Having the
dictionaries trained, for each patch of our low resolution image
we need to calculate the sparse representation. The authors of
\cite{30} used $l_1$-norm minimization method for this propose.
Here Algorithm 1 is adopted. Having the sparse
representation, calculated using the low-resolution patch, we
reconstruct the high resolution image patches using the high
resolution dictionary, $D_h$. The patches are chosen to overlap
so as to reduce the artifacts in patch boundaries. Next we
regularize and merge the patches to produce an entire image using
the reconstruction constraint \eqref{eq:SRprob}. These procedure
can be formulated as the following optimization problems:
\begin{equation}
\boldsymbol{\hat{\alpha}}_{ij} =
\underset{\boldsymbol{\alpha}}{\operatorname{argmin}} \;
\mu_{ij}||\boldsymbol{\alpha}||_0 + ||D_l\boldsymbol{\alpha} -
\mathbf{R}_{ij}\mathbf{Y}||^2_2, \label{eq:minmsealpha}
\end{equation}
\begin{equation}
\hat{\mathbf{X}} = \underset{\mathbf{X}}{\operatorname{argmin}}
\;\;  ||\mathbf{X} - \mathbf{X}_0||^2_2 + \lambda
||SH\mathbf{X}-\mathbf{Y}||^2_2, \label{eq:minmseX}
\end{equation}
where \(\mathbf{R}_{ij}\) is a matrix that extracts the (\(ij\))
block from the image, $D_l \in \mathbb{R}^{m \times n}$ is the
dictionary with \(n > m\), $\lambda$ is the regularization
parameter, $\mathbf{X}_0$ is the image obtained by averaging the
blocks obtained using sparse representation, and
\(\boldsymbol{\alpha}_{ij}\) is the sparse vector of coefficients
corresponding to the (\(ij\)) block of the image. Here,
\eqref{eq:minmsealpha} refers to the sparse coding of local image
patches with bounded prior, hence building a local model from
sparse representations. On the other hand, \eqref{eq:minmseX}
demands the proximity between the low-resolution image,
\(\mathbf{Y}\), and the output image \(\mathbf{X}\), thus
enforcing the global reconstruction constraint. In \cite{30},
$l_1$-norm minimization is used to solve \eqref{eq:minmsealpha}
, whereas we use SL0 algorithm to solve this stage. The
solution to \eqref{eq:minmseX} can be done iteratively using a gradient descent algorithm as follows:
\begin{equation}
\begin{split}
\mathbf{X}_{t+1}=\mathbf{X}_t+\nu [H^T S^T (\mathbf{Y})-
SH\mathbf{X}_t+\lambda(\mathbf{X}-\mathbf{X}_0)]
\end{split}
\label{eq:m}
\end{equation}
where $\mathbf{X}_t$ is the estimation of the high-resolution
image after the $t$-th iteration, and $\nu$ is the step size.

The proposed image super-resolution algorithm is summarized in
Algorithm 2.

\begin{algorithm}
\setlength{\leftmargini}{0pt}
\begin{enumerate}
\item \textit{Dictionary Training Phase:} train high/low resolution dictionaries \(D_l\), \(D_h\), \cite{30}\\
\item \textit{Reconstruction Phase}
\begin{itemize}
\item \textit{Sparse coding stage:} use Algorithm 1 to compute\\ the representation vectors \(\alpha_{ij}\) for
all the patches of low resolution image
\item \textit{High resolution patches reconstruction:} using the found coefficients, \(\alpha_{ij}\), the high resolution patches are reconstructed by multiplying them by\(D_{h}\)
\end{itemize}
\item \textit{Global Reconstruction:} merge high-resolution patches by averaging over the overlapped region and then use (10) to result the high resolution image.
\end{enumerate}
\caption{Image super resolution based on sparsity prior via
smoothed $l_0$-norm} \label{algo3}
\end{algorithm}
Compared to the original algorithm in \cite{30}, reduction in the
computational complexity and the possibility of improving the output
quality are expected. 

\section{EXPERIMENTAL RESULTS}
In this section we compare the proposed super-resolution
algorithm with bicubic interpolation and the method given in \cite{30}.
The image super resolution methods are tested on various images.
To be consistent with \cite{30}, patches of \(5 \times 5\) pixels
were used on the low resolution image and the scaling factor was
set to $2$. Each patch is converted to a vector with length 
$25$. The trained dictionaries, provided by authors of \cite{30},
with the sizes of $25 \times 1024$ and $100 \times 1024$ for the
low and the high resolution dictionaries were used, respectively.
To remove artifacts on the patch boundaries we set a overlap of one
pixel in the patches.

Fig. 1 and Fig. 2 (subplots (a) and (e)) depict the original Pepper and Barbara images and their corresponding
low-resolution versions. In the same figures subplots (b-d) depict reconstructed
high-resolution images using the proposed methods. Subplots (f-h) depict the corresponding SSIM maps \cite{36}. A close look on the reconstructed images, enlarged image regions (subplots (i-l)), and the corresponding SSIM maps shows that while bicubic method works pretty well in smooth regions, significant blurring occurs on  edges. The method of \cite{30} is able to recover the edges better but does not work as well in smooth regions. Compared to \cite{30} our approach is able to recover the edges and meanwhile it works better in smooth regions. One possible approach for further improving the image quality might be using a combination of bicubic and our approach.  The quantitative
results for different images reconstructed from different algorithms are shown in Table 1.
\begin{figure}[!t]
\centering \subfigure[]{
\includegraphics[width = 4cm]{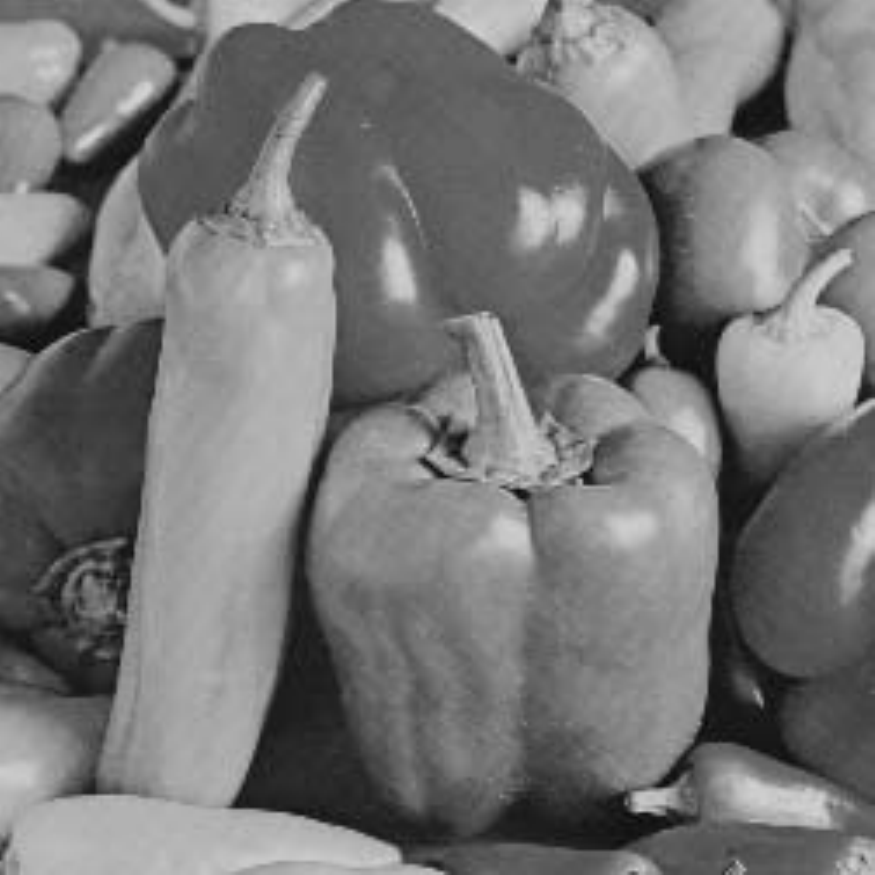}
\label{fig:tabsubfig2} }
\subfigure[]{
\includegraphics[width = 4cm]{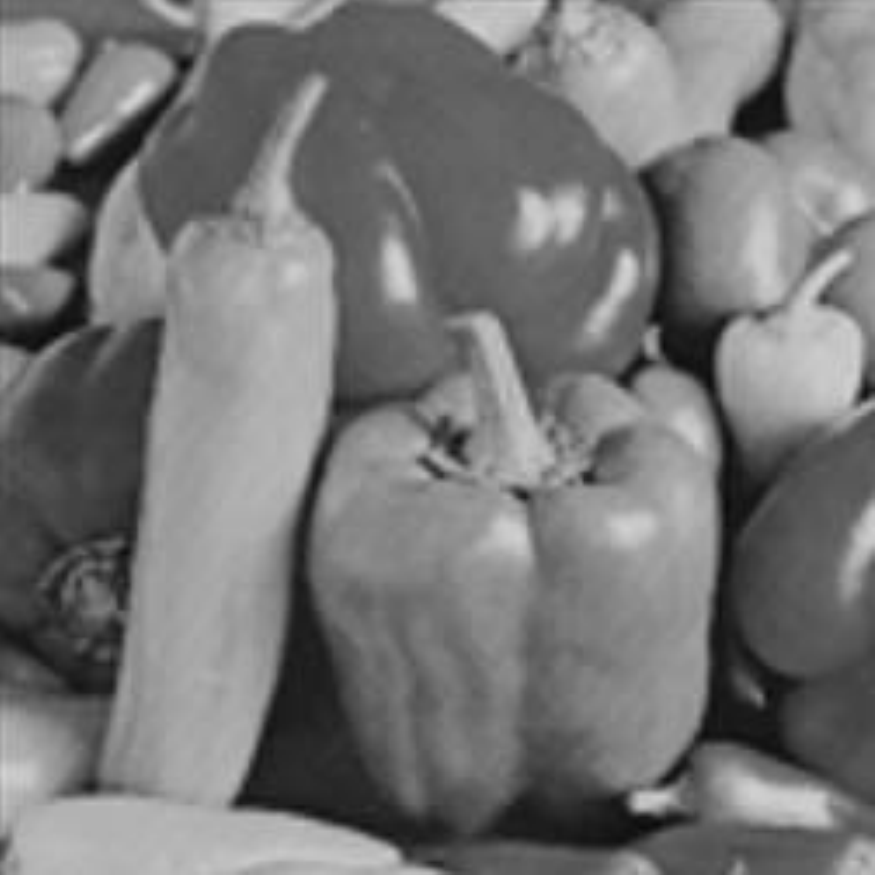}
\label{fig:tabsubfig3} }
\subfigure[]{
\includegraphics[width = 4cm]{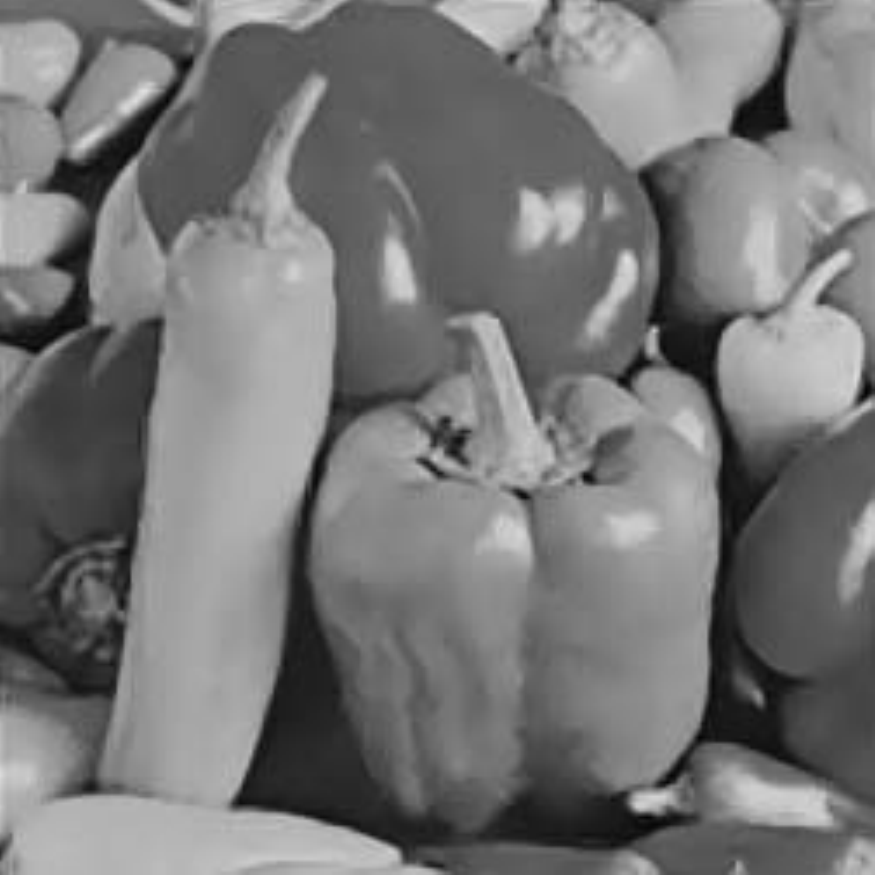}
\label{fig:tabsubfig4} }
\subfigure[]{
\includegraphics[width = 4cm]{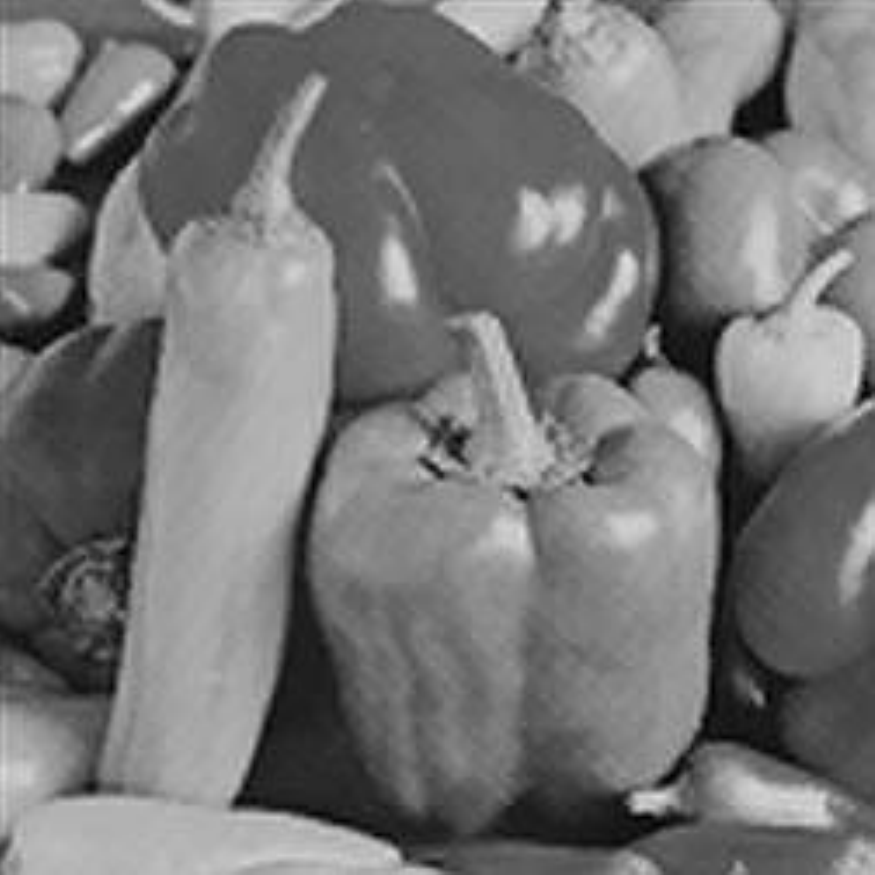}
\label{fig:tabsubfig5} }
\\
\subfigure[]{
\includegraphics[width = 2cm]{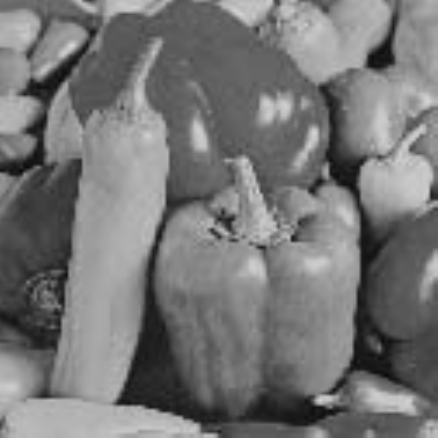}
\label{fig:tabsubfig6} }
\subfigure[]{
\includegraphics[width = 4cm]{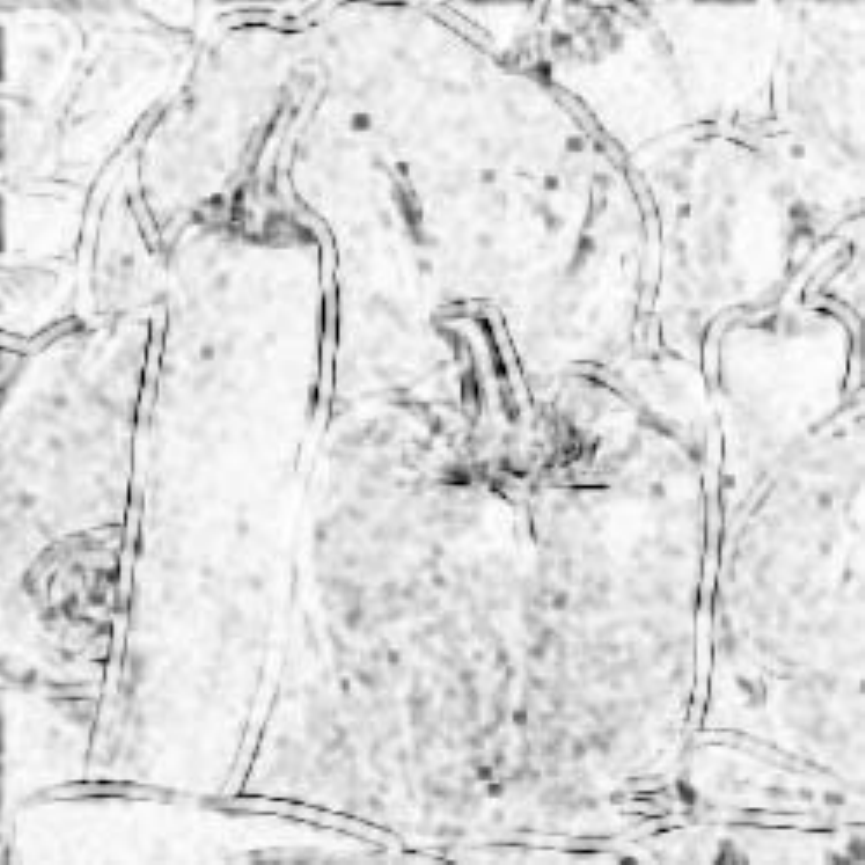}
\label{fig:tabsubfig7} }
\subfigure[]{
\includegraphics[width = 4cm]{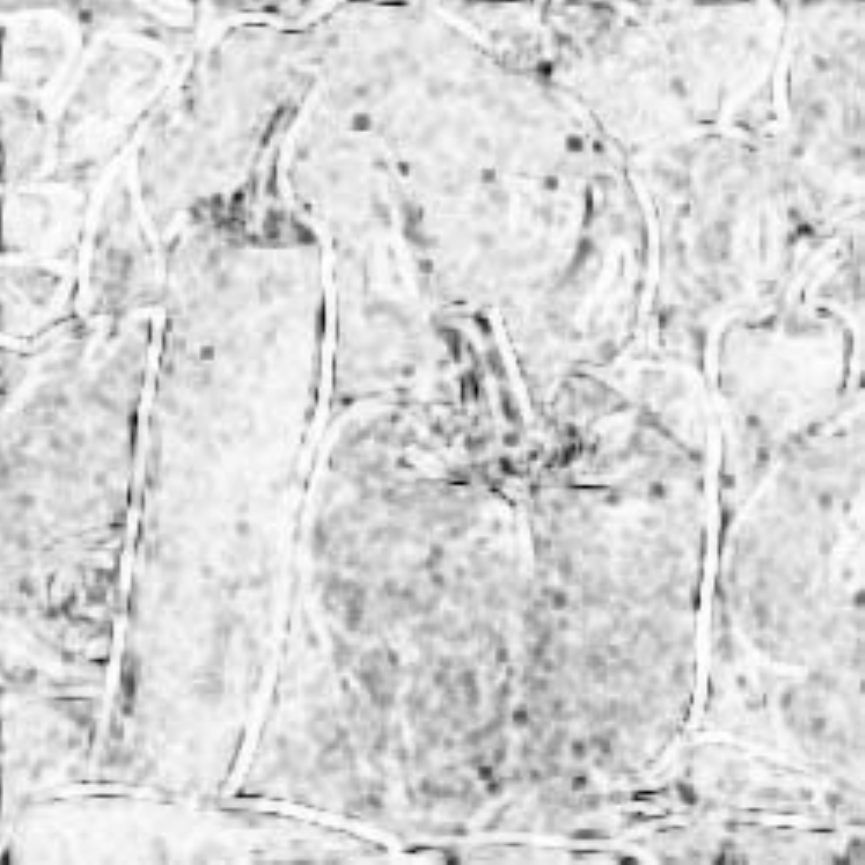}
\label{fig:tabsubfig9} }
\subfigure[]{
\includegraphics[width = 4cm]{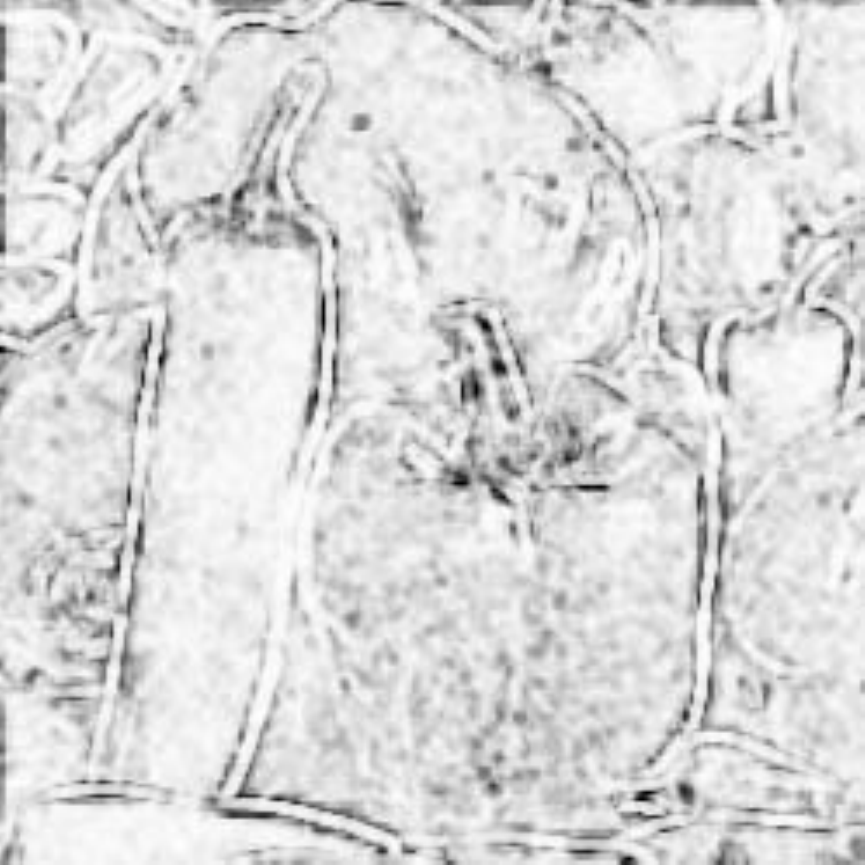}
\label{fig:tabsubfig8} }
\\
\subfigure[]{
\includegraphics[width = 4cm]{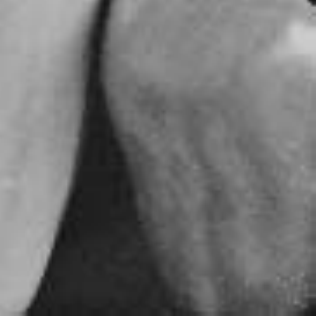}
\label{fig:tabsubfig62} }
\subfigure[]{
\includegraphics[width = 4cm]{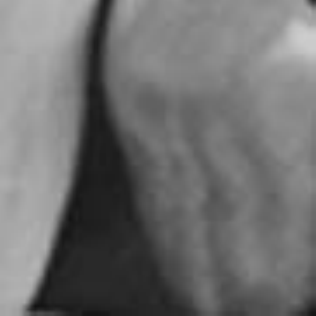}
\label{fig:tabsubfig72} }
\subfigure[]{
\includegraphics[width = 4cm]{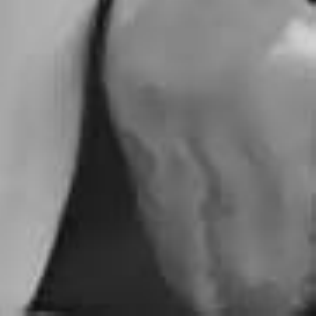}
\label{fig:tabsubfig92} }
\subfigure[]{
\includegraphics[width = 4cm]{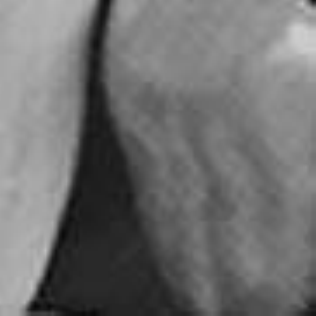}
\label{fig:tabsubfig83} }
\caption{Original Pepper image (subplot
(a)), its low-resolution version (subplot (e)), high-resolution images reconstructed by (subplot
(b)) Bicubic, (subplot
(c)) Yang et. al., (subplot
(d)) proposed method,  the corresponding SSIM maps (subplots
(f-h)), and enlarged region (subplots (i-l))..} \label{fig12}
\end{figure}

\begin{table*}[t]
\centering \caption{SSIM and PSNR comparisons of image super
resolution results} \label{tab:sresresults}
\begin{tabular}{l|cccccccccccccccc}
\hline
Image & \multicolumn{1}{c}{Barbara} & \multicolumn{1}{c}{Lena} & \multicolumn{1}{c}{Baboon} & \multicolumn{1}{c}{House} & \multicolumn{1}{c}{Watch} & \multicolumn{1}{c}{Pepper} & \multicolumn{1}{c}{Parthenon} & \multicolumn{1}{c}{Splash} & \multicolumn{1}{c}{Aeroplane} & \multicolumn{1}{c}{Tree} & \multicolumn{1}{c}{Girl} & \multicolumn{1}{c}{Bird} & \multicolumn{1}{c}{Average}\\
& \multicolumn{14}{c}{PSNR comparison (in dB)}\\
\hline
Bicubic                  & 27.08 & 32.70 & 26.34 & 33.97 & 26.94& 30.84 & 28.07 & 33.58  & 28.50 & 29.41 & 35.12 &  29.49 & 30.17 \\
Yang et al.              & 27.15 & 33.33 & 26.41 & 34.00 & 27.32& 30.41 & 28.41 & 33.87  & 29.02 & 29.60 & 34.52 &  29.63 & 30.31 \\
Proposed                 & 27.13 & 33.45 & 26.52 & 33.97 & 27.41& 30.55 & 28.68 & 34.13  & 29.24 & 29.67 & 35.25 &  29.82 & 30.49\\
\hline
& \multicolumn{14}{c}{SSIM comparison}\\
\hline
Bicubic                  & 0.744 & 0.876 & 0.700 & 0.870 & 0.790 & 0.869 & 0.763 & 0.871  & 0.857 & 0.852 & 0.925 & 0.877 & 0.832\\
Yang et al.              & 0.747 & 0.884 & 0.702 & 0.873 & 0.806 & 0.850 & 0.765 & 0.918  & 0.868 & 0.857 & 0.901 & 0.882 & 0.837\\
Proposed                 & 0.746 & 0.894 & 0.727 & 0.886 & 0.808 & 0.857 & 0.770 & 0.922  & 0.892 & 0.861 & 0.925 & 0.886 & 0.847&\\
\hline
\end{tabular}
\end{table*}

\begin{table*}[t]
\centering \caption{Execution Time comparisons of image super
resolution results (S)} \label{tab:sresresultsa}
\begin{tabular}{l|cccccccccccccccc}
\hline
Image & \multicolumn{1}{c}{Barbara} & \multicolumn{1}{c}{Lena} & \multicolumn{1}{c}{Baboon} & \multicolumn{1}{c}{House} & \multicolumn{1}{c}{Watch} & \multicolumn{1}{c}{Pepper} & \multicolumn{1}{c}{Parthenon} & \multicolumn{1}{c}{Splash} & \multicolumn{1}{c}{Aeroplane} & \multicolumn{1}{c}{Tree} & \multicolumn{1}{c}{Girl} & \multicolumn{1}{c}{Bird} & \multicolumn{1}{c}{Average}\\
& \multicolumn{14}{c}{}\\
\hline
Yang et al.              & 13.79 & 12.07 & 12.07 & 12.07 & 15.81 & 13.26 & 26.90 & 11.14  & 13.23 & 13.95 & 15.08 & 12.54 & 14.33\\
Proposed                 & 1.48  & 1.39  & 1.38  & 1.39  & 1.49  & 1.53  & 2.87  & 1.36   & 1.87  & 1.44  & 1.43  & 1.39 & 1.67&\\
Time saving        & 89.3  \% & 86.2 \% & 86.2 \% & 86.2 \% & 90.6 \% & 88.5 \% & 89.3 \% & 88.1  \% & 85.9 \% & 89.7  \%& 90.5 \% & 88.9 \%& 88.3 \%&\\
\hline
\end{tabular}
\end{table*}

All the high-resolution output images have been compared with
their original counterparts in terms of PSNR as well as of the
structural similarity index (SSIM) of \cite{36},  which is
believed to be a better indicator of perceptual image quality
\cite{100}.  It can be observed that the proposed method
outperforms the other methods in terms of both SSIM and PSNR in
most cases and on average it outperforms both methods. 

We have also included execution times for Yang et. al and our approach in Table 2. Comparison of execution times also confirms that our approach is about an order faster than the method in \cite{30}. This result is expected, SL0 is estimated to be about one to two-order faster than $l_1$-norm based minimization methods.

\begin{figure}[!t]
\centering \subfigure[]{
\includegraphics[width = 4cm]{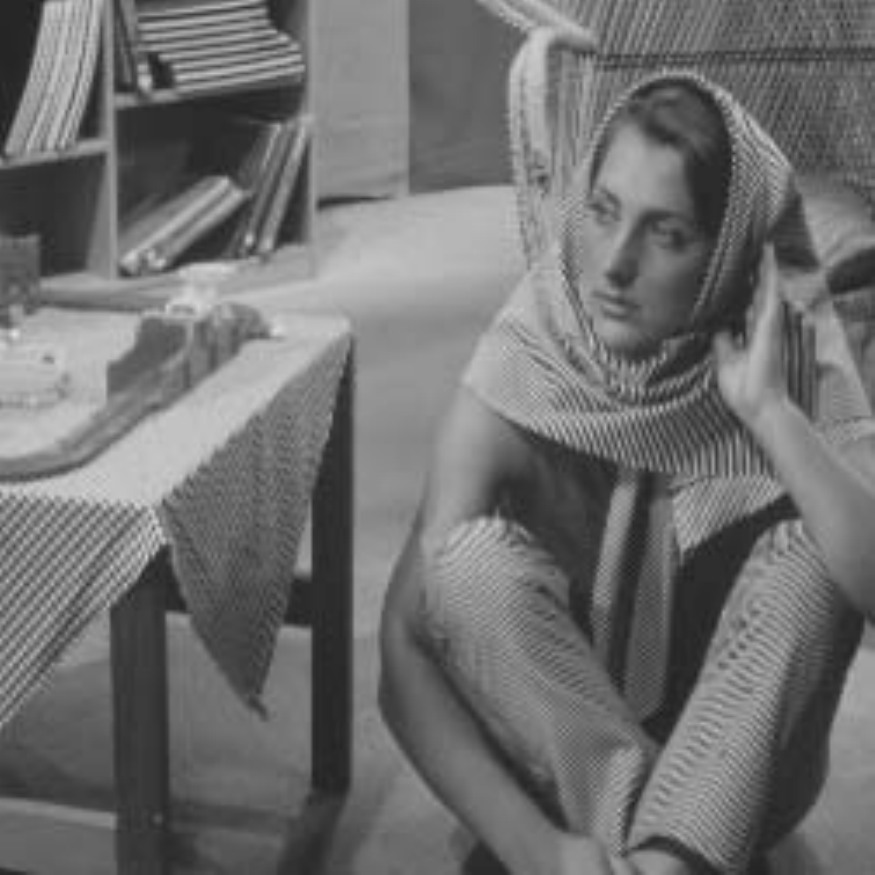}
\label{fig:tabsubfig21} }
\subfigure[]{
\includegraphics[width = 4cm]{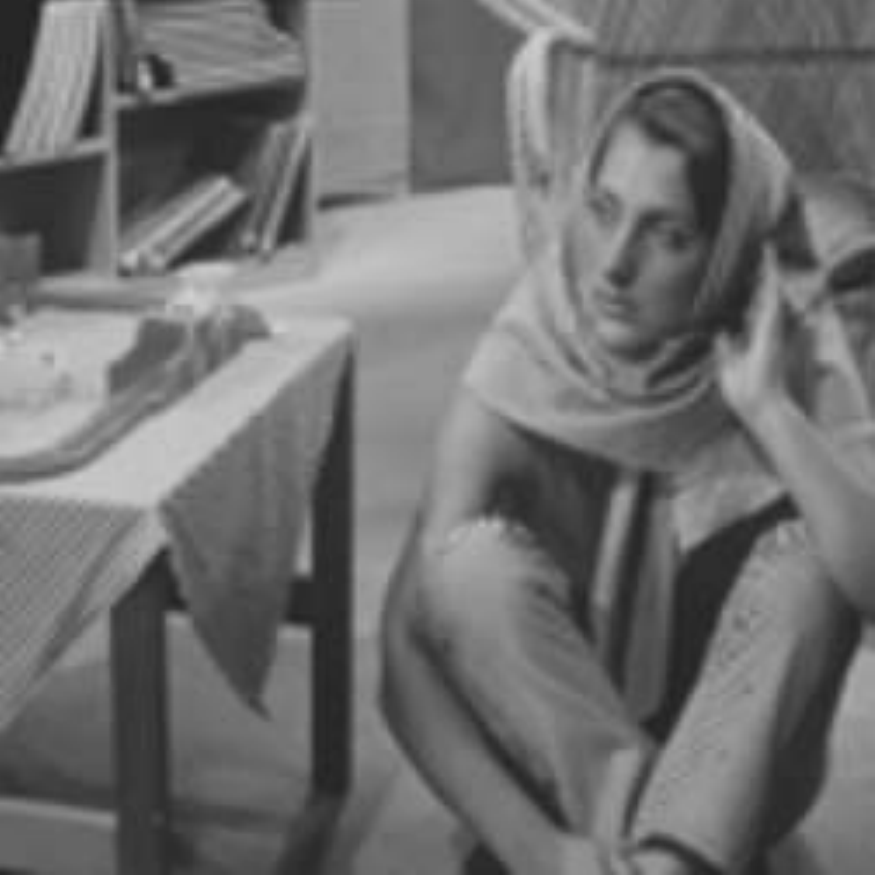}
\label{fig:tabsubfig31} }
\subfigure[]{
\includegraphics[width = 4cm]{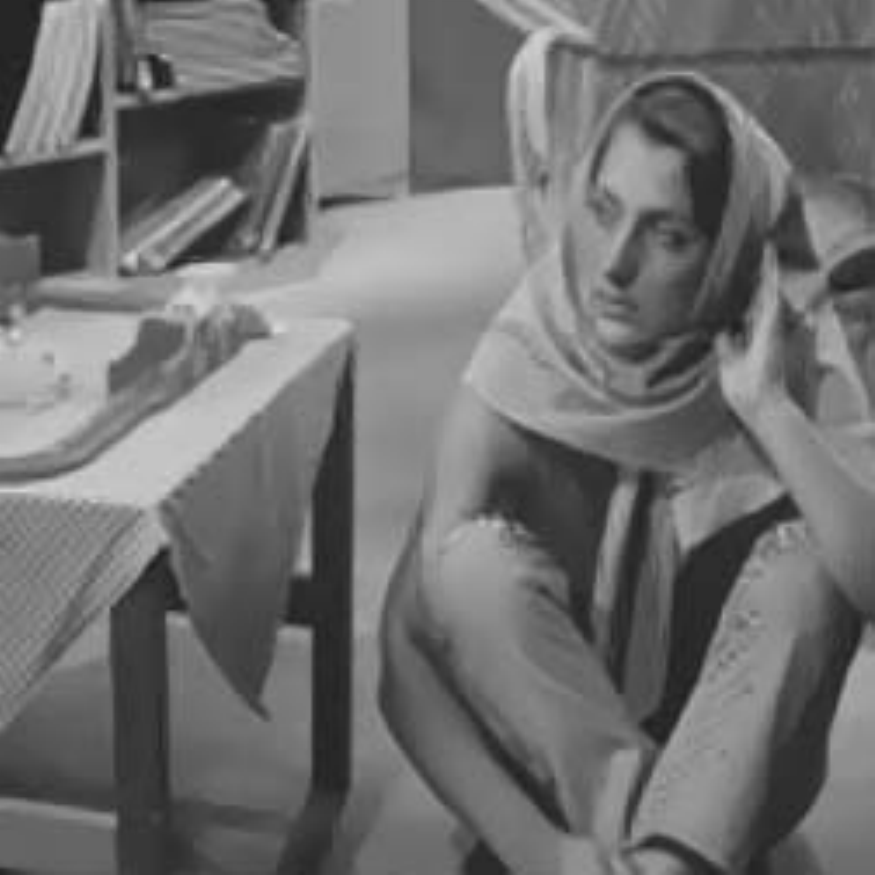}
\label{fig:tabsubfig41} }
\subfigure[]{
\includegraphics[width = 4cm]{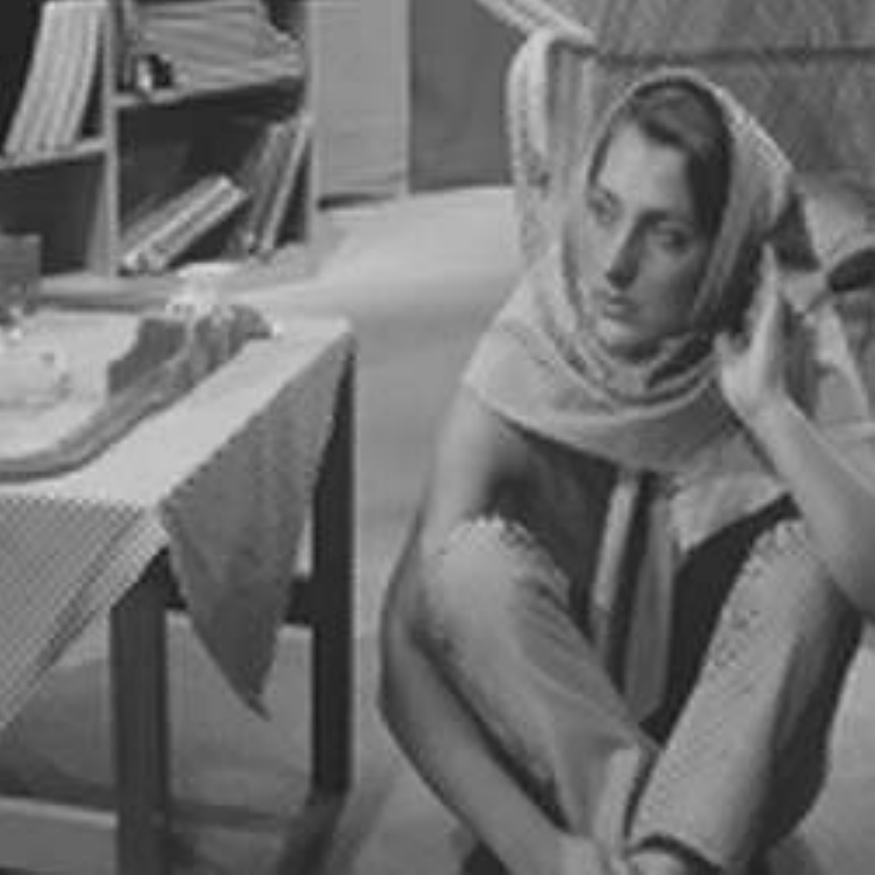}
\label{fig:tabsubfig51} }
\\
\subfigure[]{
\includegraphics[width = 2cm]{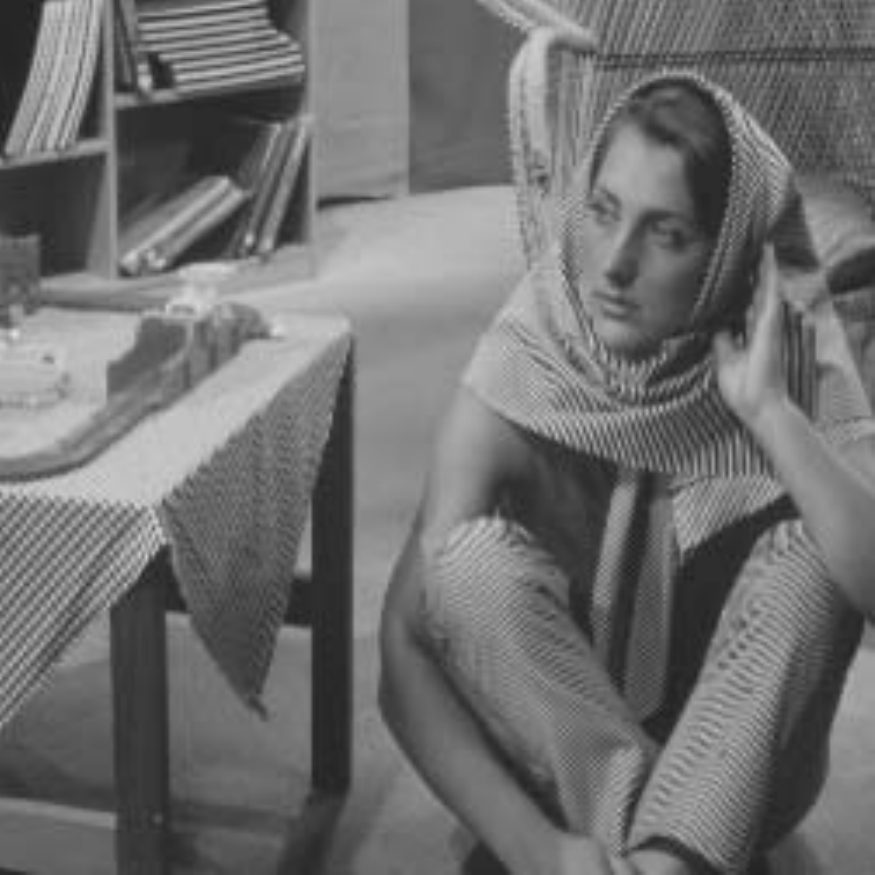}
\label{fig:tabsubfig61} }
\subfigure[]{
\includegraphics[width = 4cm]{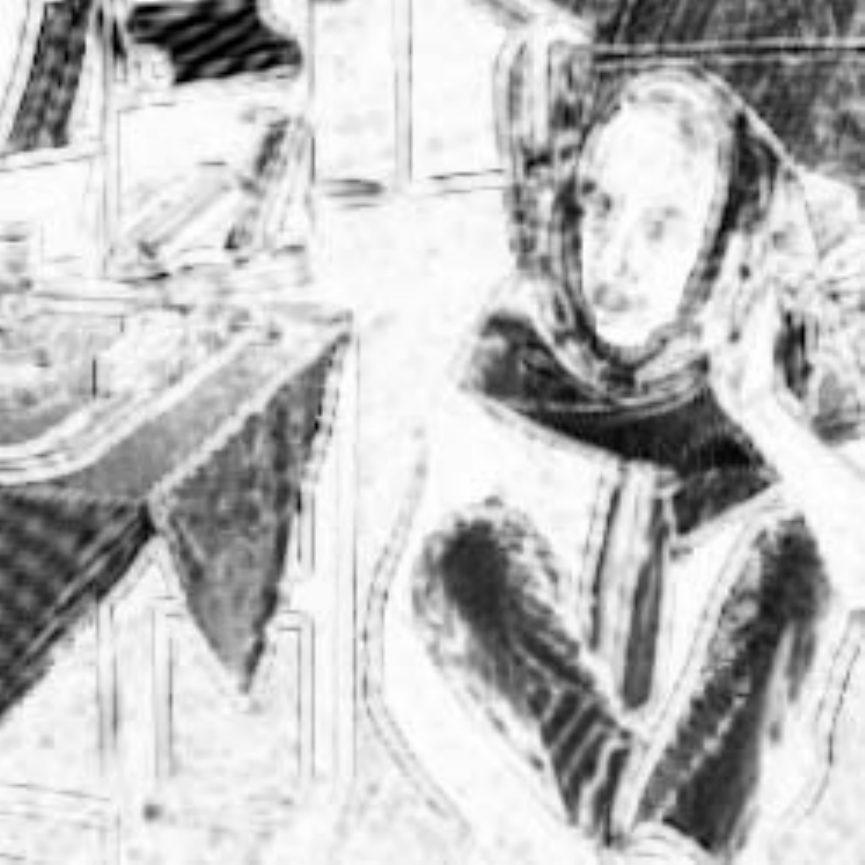}
\label{fig:tabsubfig71} }
\subfigure[]{
\includegraphics[width = 4cm]{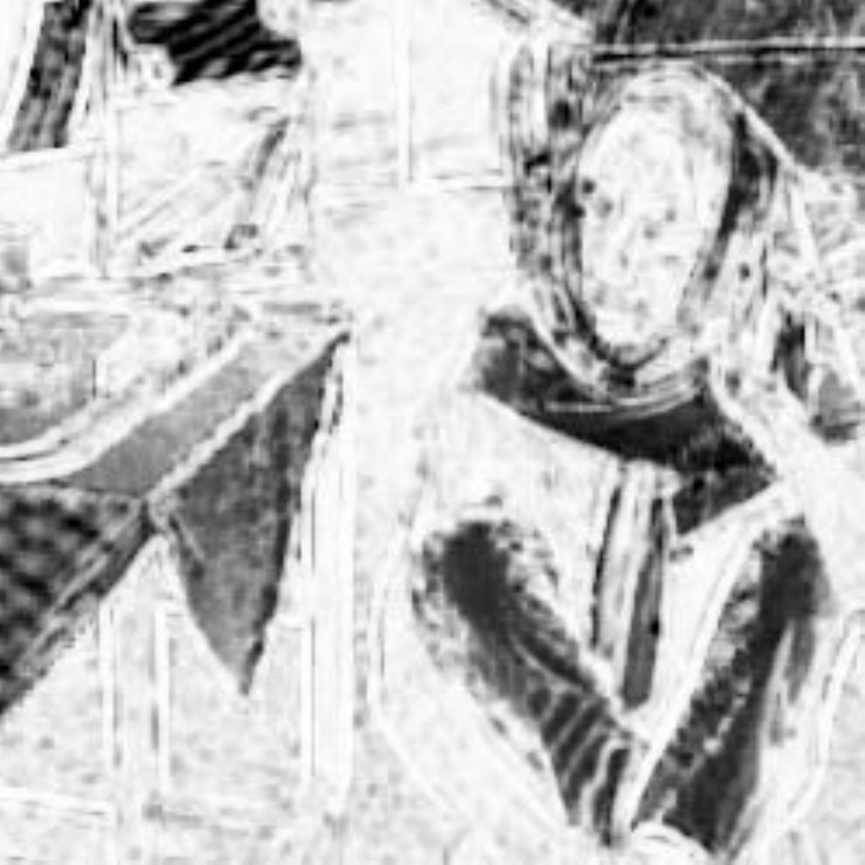}
\label{fig:tabsubfig81} }
\subfigure[]{
\includegraphics[width = 4cm]{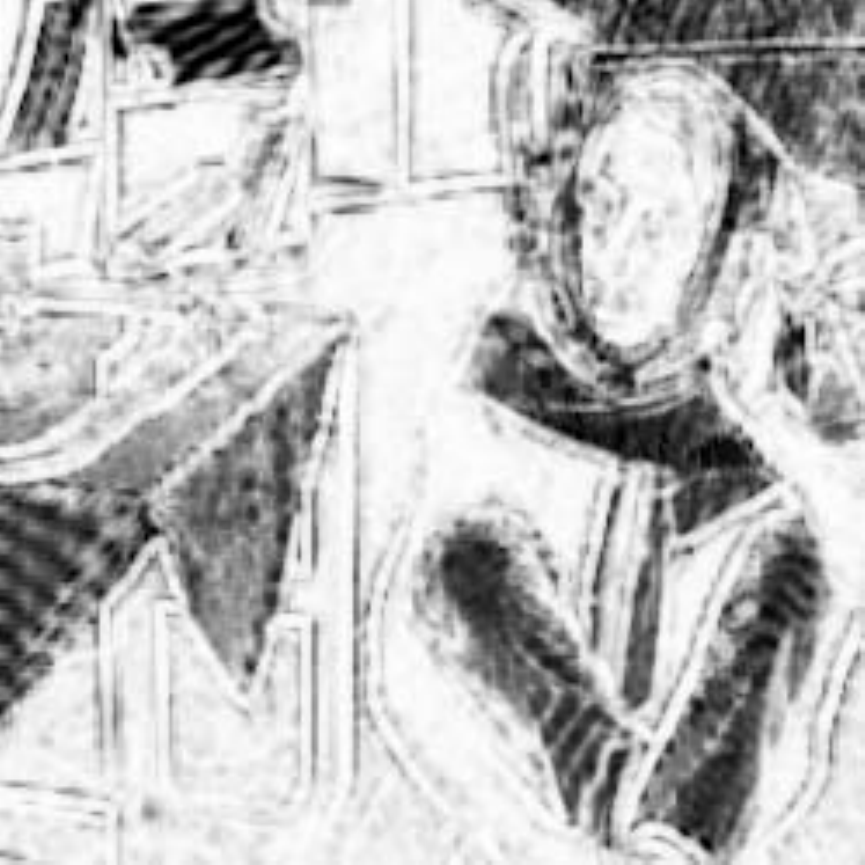}
\label{fig:tabsubfig912} }
\\
\subfigure[]{
\includegraphics[width = 4cm]{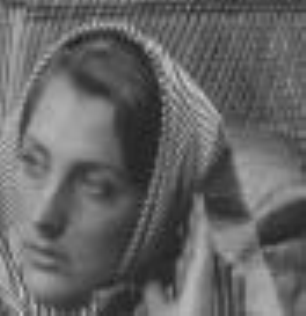}
\label{fig:tabsubfig611} }
\subfigure[]{
\includegraphics[width = 4cm]{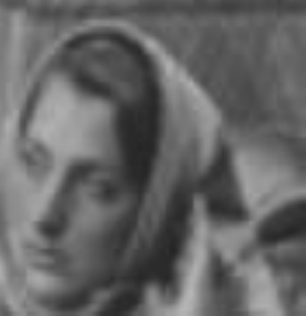}
\label{fig:tabsubfig711} }
\subfigure[]{
\includegraphics[width = 4cm]{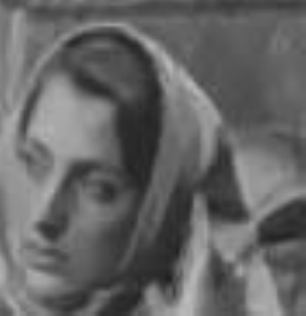}
\label{fig:tabsubfig811} }
\subfigure[]{
\includegraphics[width = 4cm]{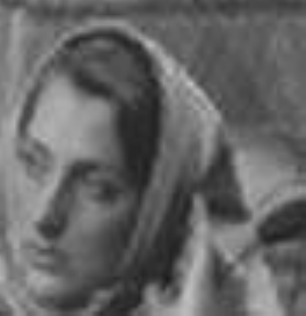}
\label{fig:tabsubfig9121} }
\caption{Original Barbara image (subplot
(a)), its low-resolution version (subplot (e)), high-resolution images reconstructed by (subplot
(b)) Bicubic, (subplot
(c)) Yang et. al., (subplot
(d)) proposed method, the corresponding SSIM maps (subplots
(f-h)), and enlarged region (subplots (i-l)).} \label{fig12}
\end{figure}
\section{CONCLUSION}
In this paper, we attempt to take advantage of the SL0 sparse coding
solver in order to improve one of the-state-of-the-art single
input image super-resolution algorithms based on sparse signal
representation.  Compared
with the method in \cite{30}, our approach significantly reduces computational
complexity, and yet improves the output perceptual quality. Our
simulations demonstrate the potential of the SL0 algorithm in improving the current image
processing algorithms that use sparse coding in one of their
stages. In the future, the algorithm maybe further improved by advanced design of the dictionary.

\section*{ACKNOWLEDGEMENT}
This work was supported in part by the Natural Sciences and
Engineering Research Council of Canada and in part by Ontario
Early Researcher Award program, which are gratefully acknowledged. The authors would also like to acknowledge Jianchao Yang for providing super-resolution codes.

\bibliography{ref}
\bibliographystyle{IEEEbib}
\end{document}